# Improving a Hybrid Graphsage Deep Network for Automatic Multi-objective Logistics Management in Supply Chain

Mehdi Khaleghi[a], Nastaran Khaleghi[b], Sobhan Sheykhivand[c], Sebelan Danishvar[d]

[a] Faculty of Industrial Engineering, Azad University of Tehran, Tehran, Iran
[b] Faculty of Electrical and Computer Engineering, University of Zanjan, Zanjan; khaleghi.nstr@tabrizu.ac.ir
[c] Department of Biomedical Engineering, University of Bonab, Bonab 55517-61167, Iran; s.sheykhivand@tabrizu.ac.ir
[d] College of Engineering, Design and Physical Sciences, Brunel University London, Uxbridge UB8 3PH, UK

## Abstract

Systematic logistics, conveyance amenities and facilities as well as warehousing information play a key role in fostering profitable development in a supply chain. The aim of transformation in industries is the improvement of the resiliency regarding the supply chain. The resiliency policies are required for companies to affect the collaboration with logistics service providers positively. The decrement of air pollutant emissions is a persistent advantage of the efficient management of logistics and transportation in supply chain. The management of shipment type is a significant factor in analyzing the sustainability of logistics and supply chain. An automatic approach to predict the shipment type, logistics delay and traffic status are required to improve the efficiency of the supply chain management. A hybrid graphsage network (H-GSN) is proposed in this paper for multi-task purpose of logistics management in a supply chain. The shipment type, shipment status, traffic status, logistics ID and logistics delay are the objectives in this article regarding three different databases including DataCo, Shipping and Smart Logistcis available on Kaggle as supply chain logistics databases. The average accuracy of 97.8% and 100% are acquired for 10 kinds of logistics ID and 3 types of traffic status prediction in Smart Logistics dataset. The average accuracy of 98.7% and 99.4% are obtained for shipment type prediction in DataCo and logistics delay in Shipping database, respectively. The evaluation metrics for different logistics scenarios confirm the efficiency of the proposed method to improve the resilience and sustainability of the supply chain.

## 1. Introduction

The processes related to managing transportation amenities construct logistics connections. Logistics play a key role in providing sustainability for the supply network. An effective logistics network helps to reduce the operation costs and brings competitive advantages to the companies. The efficiency of this network is affected by some factors, including the planning of shipment patterns and the allocation of resources. The agents, outcomes and targets affect the logistics system in the conditions related to the formation of transport facilities market. The availability and accessibility of throughput are other examples of the elements affecting the quality of transportation infrastructure. The demand fluctuations and lack of liability to the external environmental factors are examples of factors affecting the supply chain sustainability. These fluctuations lead to a variation in the pattern of transportation equipment. Hence, the predictions of the plan would not correspond to the actual variables including date of scheduled loading, type of products, number of loading trains and loading zones. Each of the mentioned subjects of the logistics process has the ability to signify the proficiency criteria. The optimization process is a compulsory element of administering the quality of logistics and management of the fluctuations as mentioned above. A positive outcome would be achieved by analyzing and opting a method with best performance.

Applying the optimization approach requires determining the goals of the logistics and the invested timelines. The existing logistics process considers the interconnection between the key elements of the administration systems of transport services and their aimed activities toward the required services. This approach facilitates the preservation and functionality of the supply network. The complete standards of obtaining efficiency in supply chain logistics are not in favor of the stock owners. These standards are not aligned with the requirements of the repository operators and the carrier owners. The aim of logistics optimization is to make the interaction of particles more perfect and satisfy the operators [1].

The accurate delivery of materials to destination and rendering the goods to the customers are the main tasks during logistics management. The main idea of logistics study is to realize an ideal balance in supply chain. The balance helps to save cost and time in an efficient supply chain. As a result, the supply chain presents services in minimum time with satisfying cost. This is the reason that the planning of optimal routes in a supply chain becomes a necessity in a sustainable supply network. The reasonable selection of the best routes has straight influence on the logistics. The automatic techniques help to promote the benefits of efficient logistics in a supply network. In the case of current business activities, the route enhancement in a chain network supports and improves the electronic commercial services. It is the fundamental element of making an intelligent transportation system. The route management of freight machines according to the intelligent procedure results in the promotion of logistics administration. The automatic approach of logistics administration reduces the traffic congestion. Besides, it helps to reduce polution emission into the environment. It is a fundamental assistance for the development of smart transportation and enhancing current electronic commercial logistics [2].

Deep learning is a new emerging approach with widespread usage in different applications of data science, signal, and image processing. It has been widely used in different scenarios of supply chain management. The approaches based on deep learning have been designed and utilized according to various stages of the supply network. The techniques of deep learning has been used in various objects of supply chain management including the planning stage for prediction of future production demand [3-5], the customer order designation in a supply chain [6], supplier selection [7, 8], the identification of suppliers [9], the transportation and delivery of the productions to the consumers [1, 10, 11], the allocation of centers for enterprises to reduce costs with appropriate choice of plant and resource location [12], production management and handling the product returns and refunds from consumers [13]. The logistics development causes the application of automatic machine learning applications in shipment and logistics operations.

Some other deep learning studies have considered risks along with the supply chain. The recognition of the threatening risks of supply chain is necessary for supply chain risk management. These risks are associated with various natural disasters, infectious disease pandemic situation, geographical vulnerabilities and financial failures. It is a significant part of supply chain administration to understand the concepts corresponding to various risks. The recognition of possible risks and the schedules for risk reduction are required for risk management. The logistics risk management is required for optimal management of a supply network.

There are some works available about for evaluation of the logistic supply chain database considering graph neural network. According to the previous studies, there is a necessity to provide a multi-task method to improve the sustainability during supply chain management. It will help to manage the logistics and related risks. Besides, it will enhance the sustainability of the supply chain network. The graph structure helps to identify the hidden connections in the database for categorization. The proposed method in this article considers the limitations of previous studies. It employs the graph structure of the data to improve the classification efficiency. Furthermore, it presents a novel geometric architecture and removes the feature selection step. It considers the original data samples for each node of graph illustration and the features are automatically extracted in each layer of the deep structure. It provides a multi-objective approach for prediction of shipment pattern, logistics ID, traffic status and logistic delay in a supply chain.

The contributions presented in this article can be introduced as follows:

(i) It provides a parallel deep network consisting of convolutional, LSTM and graphsage layers for constructing patterns to highlight differences of categories.
(ii) The suggested approach uses graph embedding of supply chain database. The hidden connections between these feature vectors are used for classification.
(iii) The proposed network architecture predicts the logistics delay.
(iv) It provides a novel structure for classification of 4 shipment types and 10 logistics ID in supply chain datasets. Also, it can identify traffic status, shipment status and logistic delays according to different datasets.
(v) It uses a parallel network of graphsage convolutional and LSTM layers for multi-objective classification of three significant supply chain databases corresponding to various scenarios.

The other sections of this paper are organized as follows. Section 2 explains recent methods of logistics and supply chain management using deep learning. In Section 3, the characteristics of the DataCo, Shipping and Smart Logistics are available in details. Besides, it covers the mathematical basis of graph attention and graphsage to describe their functionality. Section 4 unveils the principles of the proposed hybrid deep learning method according to designated targets. The modeling targets include shipment type of logistics and logistics delay prediction. Section 5 represents and extends the outcomes in terms of different evaluation metrics. The figures, tables and plots provided in this section elaborate the proficiency of the proposed approach. Section 6 is the final part assigned to conclusions.

## 2. Related Works

One of the fundamental attributes in supply chain management is transportation and logistics. These attributes certify the proficient movement of products from distributors to the end users. Optimizing the logistics and supply chain management leads to proficiency and satisfactory results. The machine learning technologies play a key role in improving the models for optimizing the management in a supply network. These algorithms pave the way for data science researchers to provide some beneficial cost-efficient programs. The application of these emerging algorithms encounters some challenges. The excessive costs of computation and complexity of data are some of the existing challenges in transportation management in a supply chain. Path planning and optimization in logistics have a positive effect on improving the delivery schedules. Improving the load management, delivery programs and route arrangement are some of the significant processes in improving the supply network internal interactions. The enhancing of logistics cost, delivery time, on-time delivery rates, thermal efficiency and economy proficiency are some of the significant items in transportation assessment in a supply chain [3]. The greenhouse gas footprint and the fuel consumptions are other emerging important factors of logistics performance in a supply chain. The restrictions of infrastructure preparation, standard conformity and alterations in demand prototypes cause difficulties in supply chain management. These difficulties make existing novelties necessary to improve logistics strategies. In this section, some deep learning strategies for solving transportation problems and supply chain logistics challenges are reviewed briefly.

Drljavca et al. studied the appearance of illegal commerce and the emergence of crime during the conflicts [10]. The fluctuations and instability of the market are the negative consequences of the insufficiencies in logistics policy for war situation and conflicts. Also, the core principal of transportation task in their study has been realized in improving a circular economy. According to their study, recycling through logistics planning plays a pivotal role in waste management.

The hydrogen supply chain has been studied by Jang et al. [14] in 2024. An algorithm has been proposed by them to model the supply chain considering the demand fluctuations and the capacity of transportation. The decision parameters resulting from the uncertainty complicate the derivation of an effective solution. The proposed methodology in their article tried to acquire an effective resolution for the stochastic programming regarding different logistics scenarios in a supply network. The low-carbon strategies has been considered in logistics management to reduce the air pollutant emissions in order to achieve green supply chain. The structural equation modeling was the basis of the algorithm by Fu et al. for green supply chain logistics [11] in 2023. They tried to link the connections in a supply chain to beneficial programs for the environment. Deng et al. in 2023 investigated a two-phase problem for producing and replenishment in the hydrogen supply chain. The pipeline failures and demand uncertainties have been the assumption in decision making in their study [15]. The decisions about replenishments were determined in response to the decisions made in the first stage and depended on service level requirements. The verification of the proposed model has been reported in terms of cumulative costs. The big data technology has been discussed for decision-making and improving a sustainable supply network by Peng et al. [16] in 2022. The combination of transportation arrangement with retailing decisions is a principal problem. The carbon regulatory policies have been considered in the study by Peng et al. for the modeling procedure. The mathematical principles have been discussed in their study and the analytics for optimization have been reported by the authors. The logistics preparation and retailing and retailing decisions have been performed through the modeling with consideration of the carbon efficient policies.

The combinations of industrial development operations have been applied for logistics management in the study by Matenga et al. [17] in 2022. These are a set of practices for automating the processes between software development and information technology operations. These strategies were based on blockchain technology. This technology is a database mechanism that allows transparent information sharing within a business network and stores data in a linked chain of blocks. They contributed to a sustainable digital economy, considering the supply chain for the railcar manufacturer. Their work resulted in a blockchain-related cloud manufacturing for producing metal parts for boxed sheets. The real time analytics of suggested method in their study showed good performance for quality control, inventory management and consumer reliability.

Niu et al. worked on capacity sharing and collaborative methodology among the producers to alleviate this type of sharing to overcome the challenges of large demand fluctuations [12]. They explored the location choice issue in logistics. The assumption in their study was about the shipping of the $3^{rd}$ party products to distribution centers, whereas the plant productions could be shipped to customer zones straightforwardly or through distribution centers. The formulation of the robust distribution model has been characterized with the use of Wasserstein sets to resolve the demand uncertainties. Sirina et al. [1] in 2021 studied the administration of cargo flows and resource allocation. The instabilities of the transport systems and demand variations affect the supply chain. Miss understanding of the internal environment and unavailability of the external principles are the deficiencies. These factors impose uncertainties to transport services. The management of freight traffic, cargo flows and the stages of process optimization have been applied to the amenities of transport services in Russian Railways. The f-score according to the results of their study has not been improved with the approach.

The choice of a competent logistics as a decision-making procedure considering multiple standards has been studied by Zulqarnain et al. [18] in 2024. The extended fuzzy sets have been used in their proposed model for interpretration of the ambiguous and unclear data. The most reliable logistics company has endorsed and showcased the credibility of the suggested decision-making transport technique for sustainable supply chain. Einstein type of aggregation operators have been considered in their study, however the accuracy level has not been improved by their approach. The supply chain responsiveness in developing countries has been assessed by Moh'd Anwer [19] in 2022 corresponding to logistics strategy. Their study examined the effects of delivery expedition on the connection between logistics plans and supply chain efficiency. The dataset in their study consisted of 212 participants in large manufacturing firms in the Middle East Territory. The impacts of supply chain sustainability on the performance of production firm have been examined in their study. The conceptual model proposed by them has shown the effect of delivery reliability on the relation between logistics and supply chain. Improving the performance of the supply chain has been achieved in response to their model.

Wang et al. presented a deep analytics for the investigation of transport capacity deficiencies considering Australian logistics service providers. All transport types encounters driver insufficiency and the study by Wang et al. investigated the connections between the logistics deficiencies and logistics performance [20]. The structural equation modeling has been utilized in their study. Automotive parts producing plants require a network of transportation. Infrastructural impediment and insufficient elaborate workers are the significant disruption factors in logistics network of the automation industry. The transportation disturbance components of automotive parts manufacture enterprises have been analyzed in the study by Fartaj et al. [21] in 2020. The best-worst and strength-relation analytical method framework has been tested on data from a company in Oldcastle located in Canada. In the study by Tsolaki et al. [22] in 2023, the arrival time, demand forecast, traffic location and congestion prediction, the vehicle routing problem and disturbance detection in supply chain logistics data have been reported as a review study. The methods according to machine learning algorithms have been classified to demonstrate the methods' evolution in recent years for different applications.

In contrast to the past tendency of nations for globalization in producing goods, some nations have a desire for domestic manufacturing. The beneficial factors of incremental security, improved local jobs and developed economies are the results of the increased domestic manufacturing. The effects of structuring a supply chain framework and the expansion of domestic production have been studied by Chen et al. [23] in 2023. The social costs have been assessed during the study. They investigated the impact on the functionality and preservation of a nation's roadways. The traffic congestions, preservation costs, fuel consumption, pollutant emissions and traffic incidents have been recognized as the effects of expanding the domestic productions. The concepts of the mathematical modeling of multi-echelon, multi-commodity location issues have been exploited for a case study of U.S. domestic production of N95 filtering face mask respirator. The prediction of the noticeable impacts on the local raw materials and middle products transportation was true and it has been realized during the study. The estimation of domestic manufacture capability has been acquired in terms of household truck loads and motor vehicle miles during the study.

As stated in the study by Xiaodi Xu [2] about logistics management, Fenglin et al. proposed a deep reinforcement learning method for a particular cold chain logistics. The global route planning has been performed through the upper layer. The local temperature control regulations has been performed within the lower part of the network. The suggested method caused an improvement of cargo loss and the reduction in energy consumption. The application of DQN network has been explored by Guo Canbo [2] for urban express delivery. The dynamic path adjustment has been realized in their study by constructing a state-action mapping table. The weather factors and traffic congestion have been incorporated innovatively into the state space modeling. The model has been applied to a logistics center in Beijing and improved the distribution efficiency. A rural route optimization approach has been proposed by Xin Rongyan et al. [2] based on Policy Gradient. The study designed model with the aim of shortest distribution path. The distribution cost and constraints have been included in their model and Monte Carlo tree search in order to generate the methodology. The traditional genetic algorithm reduced the number of iterations and distribution cost. Their method converged to the optimal solution with a lower number of iterations.

A criticism method based on dominant actor has been studied by Zhao Yan [24] for delivery scenarios. The AlphaGo's Monte Carlo Tree combined with reinforcement learning has been applied to vehicle planning for the first time. The convolutional layers have been used in the proposed model to employ for the verification of the effectiveness on the Amazon transportation dataset. The urban transportation network proposed by the authors has been applied to New York City's logistics network. The number of empty cars and delivery delays decreased. Shortening the distribution time corresponding to transportation network optimization in rural areas has been solved using a meta-reinforcement learning distribution model [2] by Xiale et al. in 2024. The path optimization effect has increased in the transportation evaluation in Southeast Asia employing their proposed road network. There were many challenges including the algorithm computational burden. The risk of data leakage existed in multi-agent collaboration, the long-term credit allocation problem in dynamic environment and delayed reward leaded to fuzzy strategy update solution.

Remotely Piloted Aircraft Systems (RPAS) are one of the studies of transportation technologies attracted attentions in recent years. The transportation services have been provided to customers to serve them in a collaborative logistics network. The low capital and functionality costs are the advantages of RPAS. Having the ability to landing and takeoff with a cost-efficient infrastructure in a restricted space, the essential existence of an onboard pilot has been eliminated. In this way and considering the advantages as mentioned earlier, RPAS are beneficial methods for transportation activities. The energy consumption and the air pollution would be reduced as a result of efficient delivery processes [25-28].

A simulation of logistics assessment has been performed by Diaz et al. in order to investigate the logistics levers of pharmaceutical manufacturing firms. In their study related to pharmaceutical transportation, they assessed the uncertainty impact of logistics in global supply chain [29, 30].

Sustainable Halal logistics in business operations has been implemented in business operations. The collaborations with halal stakeholders is beneficial for sustainability of a business. Overseeing the segregation of permissible halal products from non-permissible ones is an important factor to be considered in a sustainable halal supply chain. Some studies have investigated halal transportation modeling in order to achieve this aim. The permissible products must be clean and safe. Non-halal logistic providers should not be utilized for carrying halal materials. The probable contamination should be prevented to achieve sustainable aims [31, 32]. Shariah standard halal logistics has been constructed the criterion for separation of the permissible from non-halal products. A halal brand identity has been used in order to adapt to the competitive domain [32]. In order to obtain a halal brand identity, it is required to adapt to a competing environment via a constructive halal logistic method for supply network [33]. Due to the sizeable market demand of Malaysia, the logistics supply chain industry attempted to transform the industry and aimed to adhere them to halal logistics' specifications [34]. Halal logistic supplier providers require standards, technology, logistics awareness and transportation knowledge to stay competitive in halal transport providers [34].

In this article, we propose a novel hybrid geometric network for automatic logistics supply chain management. In the next section, we explain the database settings and the mathematical background of our proposed method.

3. **Materials and Methods**

In this section, the details of the three databases used in this study are explicated. The DataCo, the Shipping and the Smart Logistics databases are used in this study. Besides, the mathematical basics of graphsage and graph attention network will be elucidated to understand how the graph layers work in a graph deep network.

**3.1. Database setting:**

Table 1 illustrates the details of the DataCo dataset. A set of 36000 transactions of DataCo global company have been analyzed to cover the 4 types of shipping mode in logistics. The types of transaction, days for shipment (scheduled), days for shipping, benefit per order, sales per customer, latitude, longitude, order item discount rate, order item discount, order item total and order profit per order are the characteristics for each data sample. The target labels for these data samples are the late delivery risk status. Besides, the four types of shipping mode considering standard, first class, second class and same day would be classified with the four-category classification model.

Table1 . DataCo specifications.

| DataCo | Feature | Format |
|---|---|---|
| 1 | Type | (4 different types of word ) Debit-0, Transfer-1, Payment-2, Cash-3 |
| 2 | Real days of shipping | Digit |
| 3 | Planned days of shipment | Digit |
| 4 | Gain for customer order | Numeral |
| 5 | Sales for cunsumer | Numeral |
| 6 | Latitude of location | Numeral |
| 7 | Longitude of location | Numeral |
| 8 | Discount | Numeral |
| 9 | Discount rate | Numeral |
| 10 | Total order | Numeral |
| 11 | Rate of order profit | Numeral |
| 12 | Order state | (8 different text type) 0-Complete,1-Processing, 2-PendingPayment, 3-Closed,4-Pending, 5-On-hold,6-Suspected-fraud, 7-Canceled, 8-Payment-Review |

Table 2. Target tasks for DataCo.

| DataCo | Target Feature | Explanation |
|---|---|---|
| 1 | Delivery status | 2-Late delivery 1-On-time |
| 2 | Shipping mode | 4- Same Day 3-Second Class 2-First Class 1-Standar Class |

The Shipping database is the second dataset utilized in this research. A set of 5280 transfers in this database has been analyzed in this article. The specifications and targets of this database are available in Tables 3 and 4, respectively. There are 5 warehouses and there are 3 different categories for shipment mode and 2 for reached time classification.

The third one is the Smart Logistics database. A set of 1000 transfers in this database is investigated in this article. The principles about this dataset are accessible in Tables 5 and 6, respectively. There are 10 IDs, 3 types of shipment status, 3 types of traffic status and 2 categories of logistics delay for classification.

Table 3 . Shipping Dataset Features.

| Shipping | Feature | Explanation |
| --- | --- | --- |
| 1 | Customer care calls | Digit |
| 2 | Customer rating | Digit |
| 3 | Cost of the product | Numeral |
| 4 | Prior purchases | Digit |
| 5 | Product importance | 3 types of word (low, medium, high) |
| 6 | Gender | 2 types of word (F,M) |
| 7 | Discount offered | Numeral |
| 8 | Weight in grams | Numeral |

Table 4 . Target tasks in Shipping dataset.

| Shipping | Target Feature | Explanation |
| --- | --- | --- |
| 1 | Warehouse | 5 categories ( A, B, C, D, F ) |
| 2 | Mode of Shipment | 3 categories ( Flight, Ship, Road ) |
| 3 | Reached On-Time | 2 categories ( Late, On-Time ) |

Figure 1 illustrates the DataCo characteristic signals for all transactions in the company.

It shows the fluctuations related with specific characteristics of DataCo including 'order item discount', 'sales per consumer', 'longitude of location', 'total order item'.

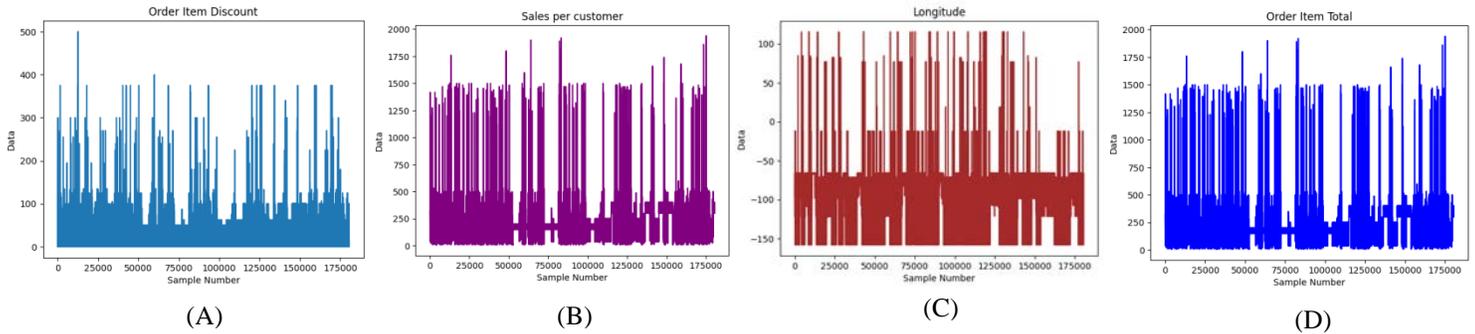

(A)        (B)        (C)        (D)

Figure1 . Characteristic plots for Dataco dataset. (A) Order discount (B) Sales (C) Longitude of location (D) Order item total.

Table 5 . Smart Logistics Dataset Features.

| Smart Logistics | Feature | Explanation |
|---|---|---|
| 1 | Latitude | Numeral |
| 2 | Longitude | Numeral |
| 3 | Inventory_Level | Numeral |
| 4 | Shipment_Status | 3 categories (Delayed, In Transit, Delivered) |
| 5 | Temperature | Numeral |
| 6 | Humidity | Numeral |
| 7 | Traffic Status | 3 categories (Detour, Heavy, Clear) |
| 8 | Waiting Time | Numeral |
| 9 | User Transaction Amount | Numeral |
| 10 | User Purchase Frequency | Digit |

Table 6. Target tasks in Smart Logistics dataset.

| Smart Logistic | Target Feature | Explanation |
|---|---|---|
| 1 | Truck_ID | 10 categories ( Truck_1, Truck_2, Truck_3, Truck_4, Truck_5, Truck_6, Truck_7, Truck_8, Truck_9, Truck_10 ) |
| 2 | Shipment Status | 3 categories ( Delayed, In Transit, Delivered ) |
| 3 | Traffic Status | 3 categories ( Detour, Heavy, Clear ) |
| 4 | Logistics Delay | 2 categories ( Late, On-Time ) |

### 3.2. Graphsage Formulation

The embedding generation part of the Graphsage algorithm is described in this section. The denoted variables of K aggregator functions are learned during the training stage. The aggregated information from node neighbors as well as set of weight matrices is utilized to propagate information through the layers of the search depths.

Algorithm 1: The pseudo code for GraphSage.

*Input*:
*Graph* $G(v(nodes), e(edges))$; *input feature vectors* $\{x_v\}$; *depth* $K$; *weight matrices* $W^k$;
$\Delta$ *nonlinear function*; *Aggregate* $_k$, $\forall k \in \{1,...,K\}$; *neighboring function* $NF: v \to 2^v$
*Output*:
$z_v$ (*representation vector*)
$x_v \to h_v^0$
*for* $k = 1,...,K$ *do*
*for* $v \in V$ *do*
$Aggregate_k(\{h_i^{k-1}, \forall i \in NF(v)\}) \to h_{NF(v)}^k$;
$\Delta(W^k \cdot Concatenation(h_v^{k-1}, h_{NF(v)}^k)) \to h_v^k$
*end*
$h_v^k / \|h_v^k\|_2 \to h_i^{k-1}$
*end*
$h_v^K \to z_v$

The embedding needs to aggregate information from the representations of the nodes in its neighborhood into a single vector such as $h_{NF(v)}^{k-1}$. The concatenation of the node's representation with the aggregated vector is performed through the next step. The fully connected layer is used in this stage.

Learning the parameters of the aggregators via stochastic gradient descent is performed using a graph-related loss function to construct the output representations. Tuning the weight matrices is the next step. The graph-based cost function encourages neighboring nodes to have similar representations, while imposing a distinctive representations to the contrasting nodes.

$$J_G(z_u) = -\log(\Delta(z_u^T z_v)) - Q.E_{v_n \sim P_n(v)} \log(\Delta(-z_u^T z_{v_n})) \qquad (1)$$

In this formula, v co-occurs near u on fixed-length random walk, $\Delta$ is the sigmoid function. $P_n$ is the distribution function for negative sampling, and Q demonstrates the number of negative samples. The $z_u$ is the representation vector generated from node's local neighbors. The unsupervised loss in the above equation can be replaced by a task-specific goal.

There is no natural ordering in a node's neighbors. The aggregator functions in the graphsage algorithm must be able to operate over an unordered collection of vectors. It would be symmetric while still being trainable. The symmetrical property of the aggregator certifies that the model can be employed to randomly order neighbor feature sets of a node. Three kinds of aggregator functions have been examined including mean, LSTM and pooling aggregators.

The mean aggregator employs the mean operator. The mean operator is the first candidate aggregator function, and the element-wise mean of the neighboring feature vectors should be considered for the aggregator. This function is similar to the convolutional propagation in the graph convolutional network. A variant of the graph convolutional network method

$$\Delta(W.Mean(h_v^{k-1} \cup \{h_j^k, \forall j \in NF(v)\})) \to h_v^k \qquad (2)$$

This is the customized mean-based aggregator convolutional and it is a linear estimation of a localized spectral convolution.

A more complex function for aggregation in graphsage modeling is the one with LSTM structure. There is a larger expressive capability for this type of structure. They do not contain symmetric structure and they process the inputs in a sequential mode. The adaptation of LSTMs to operate on an unordered set of neighbors is performed by applying the LSTMs to a random set of the neighbors.

The pooling method is another type of aggregation. It is symmetric and it can be trained considering each neighbor's feature vector. These features are fed using a dense layer and a max-pooling is applied in order to aggregate the information. By employing this operator, various aspects of the neighboring set would be captured. Other symmetric functions for example an element-wise mean function can be used instead of max in this formula.

$$Aggregate_k^{pool} = \max(\{\Delta(W_{pool} \, h_{w_j}^k + b), \forall w_j \in NF(v)\}) \tag{3}$$

### 3.3. Graph attention

This section describes the formulation of the graph attention layer. A set of features as the input of the graph attention layer, N and F designate the number of nodes and features, respectively.

$$f = \{\vec{f}_1, \vec{f}_2, ..., \vec{f}_N\}, \vec{f}_i \in R^F \tag{4}$$

A new set of node features would be created as the output of the graph attentional layer.

$$f' = \{\vec{f}'_1, \vec{f}'_2, ..., \vec{f}'_N\}, \vec{f}'_i \in R^{F'} \tag{5}$$

The weight matrix $W \in R^{F' \times F}$ has been applied to every single node. The mechanism of self-attention is employed to calculate the attention coefficients :

$$attention: R^{F'} \times R^{F'} \to R \quad a_{mn} = attention(W \vec{f}_m, W \vec{f}_n)$$

A leaky rectified linear unit can be employed to calculate the normalized output considering a non-linear activation function.

$$a_{mn} = \frac{\exp(Leaky \, ReLU(\vec{w}^T [concatenation(W\vec{f}_m, W\vec{f}_n)))}{\sum_{k \in N_i} \exp(Leaky \, ReLU(\vec{w}^T [concatenation(W\vec{f}_m, W\vec{f}_k)))} \tag{6}$$

Considering the first-order neighboring nodes, the normalization process is performed across all choices of j using the softmax function:

$$sa_{mn} = soft\max_n(a_{mn}) = \frac{\exp(a_{mn})}{\sum_{k \in N_i} \exp(e_{ik})} \tag{7}$$

The normalized attention coefficients are considered and a nonlinearity is imposed to the output.

$$\vec{f}'_m = \Delta(\sum_{n \in N_m} sa_{mn} W \vec{f}_n) \tag{8}$$

The concatenation of the feature is required to construct the output.

$$\vec{f}'_m = \underset{k=1}{\overset{K}{Concatenation}} \Delta(\sum_{n \in N_m} sa_{mn}^k W^k \vec{f}_n) \qquad (9)$$

For a multi-head attention on the final layer of the network for prediction, the averaging should be employed and the final classification layer should be considered after the averaging step.

$$\vec{f}'_m = \Delta(\sum_{k=1}^{K} \sum_{n \in N_m} sa_{mn}^k W^k \vec{f}_n) \qquad (10)$$

## 4. Methodology

The graphical diagram of different stages in accordance with the proposed method is represented in Figure 2. The Dataco, Shipping and Smart Logistics datasets are used in this study. As it can be seen in this figure, after the pre-processing of the data and graph design stage, the acquired graph would be applied to tune the parameters of the proposed hybrid graphsage network (H-GSN) during the training stage. The network includes three distinct parts of deep networks. The graph-based section consists of four sequential layers of GraphSage kernel. The convolutional part includes two sequential non-graph kernel layers and the LSTM consists of sequential layers. The loss function of the hybrid-graph network is the weighted summation of the parallel geometric part, convolutional part of the network and the LSTM section. The training phase of the H-GSN is performed with K-fold cross-validation.

### 4.1. Pre-processing stage:

The Dataco, Shipping and Smart Logistics datasets are considered in this study. The conversions of text-like features in datasets to integers are the first step of the pre-processing stage. The selection of features in order to clean them is another step. The clean array of features is used and imposed to the graph design phase. Cleaning the data to have a balance between different categories during classification is the next step in this stage.

The target for training the proposed H-GSN is considered as the zero-one conversion of the on-time delivery status and late delivery into zero and one, respectively. Also, the digit conversion of the shipment type has been considered for this database. For the Shipping database, the target is the shipment mode, warehouse number and binary digit conversion of on-time reaching. The targets regarding the Smart Logistics dataset are logistic IDs, digit conversion of traffic status and shipment status.

Standard scaling is the next part of the pre-processing. It is the necessary part for an optimal training of our proposed method. Windowing for constructing graph of neighboring nodes is another step of the pre-processing stage.

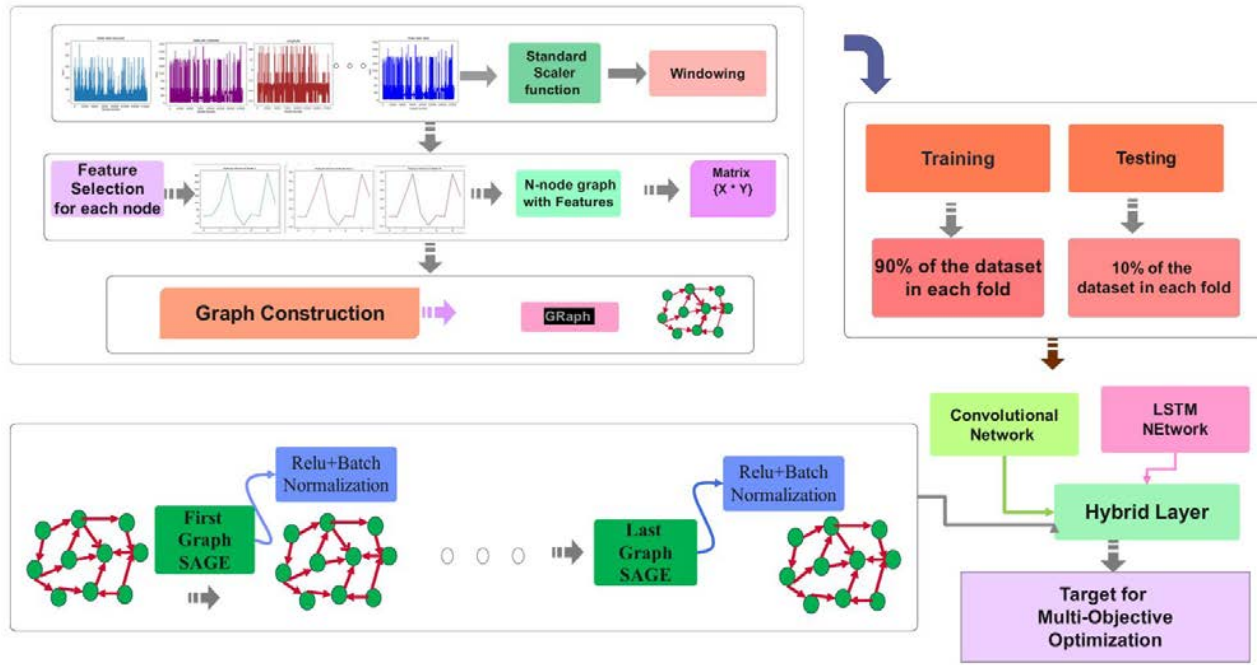

Figure 2. The schematic overview of the suggested H-GSN.

## 4.2. Graph construction:

After pre-processing, the graph design stage is necessary to employ the acquired graph in the training phase of the proposed network architecture. The correlation of characteristic features in transaction data in three databases is required for graph embedding. A rectified leaky unit is utilized for computing the absolute value of the cross-correlation matrix. Also, a threshold level is necessary to clean the output array and decrease the computational burden of the algorithm. The adjacency matrix is the output of the leaky rectified linear unit and threshold stage according to the simplified graphical representation of the graph design stage in Figure 3.

## 4.3. Proposed H-GSN architecture:

Figure 4 delineates the detailed graphical representation of the proposed network architecture. As this figure shows, our proposed geometric H-GSN contains four layers of graph convolution. As specified by this figure, in every GraphSage layer, the first step is the estimation of the GraphSage of the input graph. The next layer is the activation layer. Also, batch normalization is utilized in the output of each layer to normalize the input to the next layer.

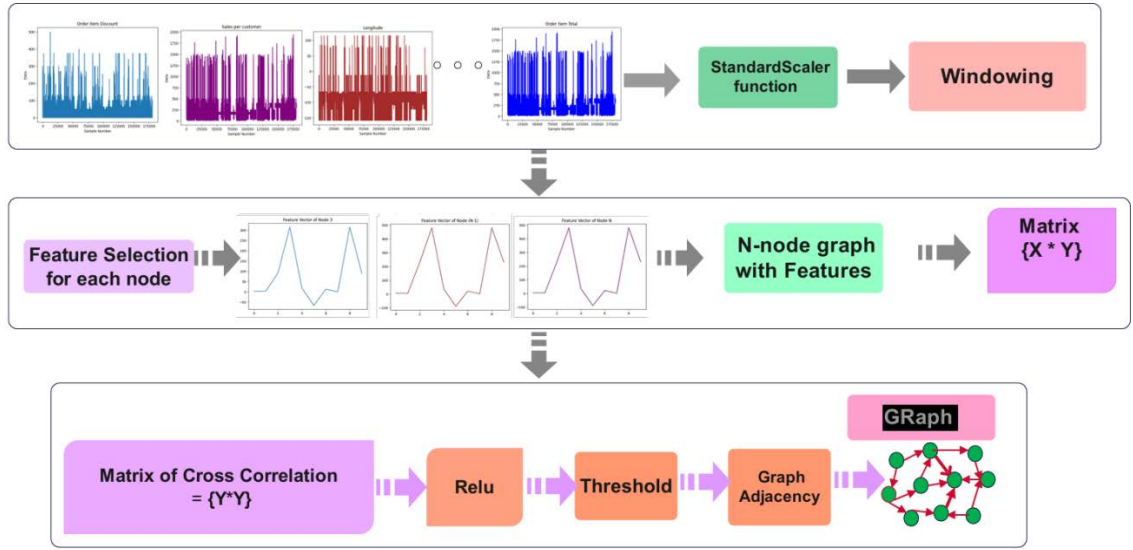

Figure 3. Graph construction stage.

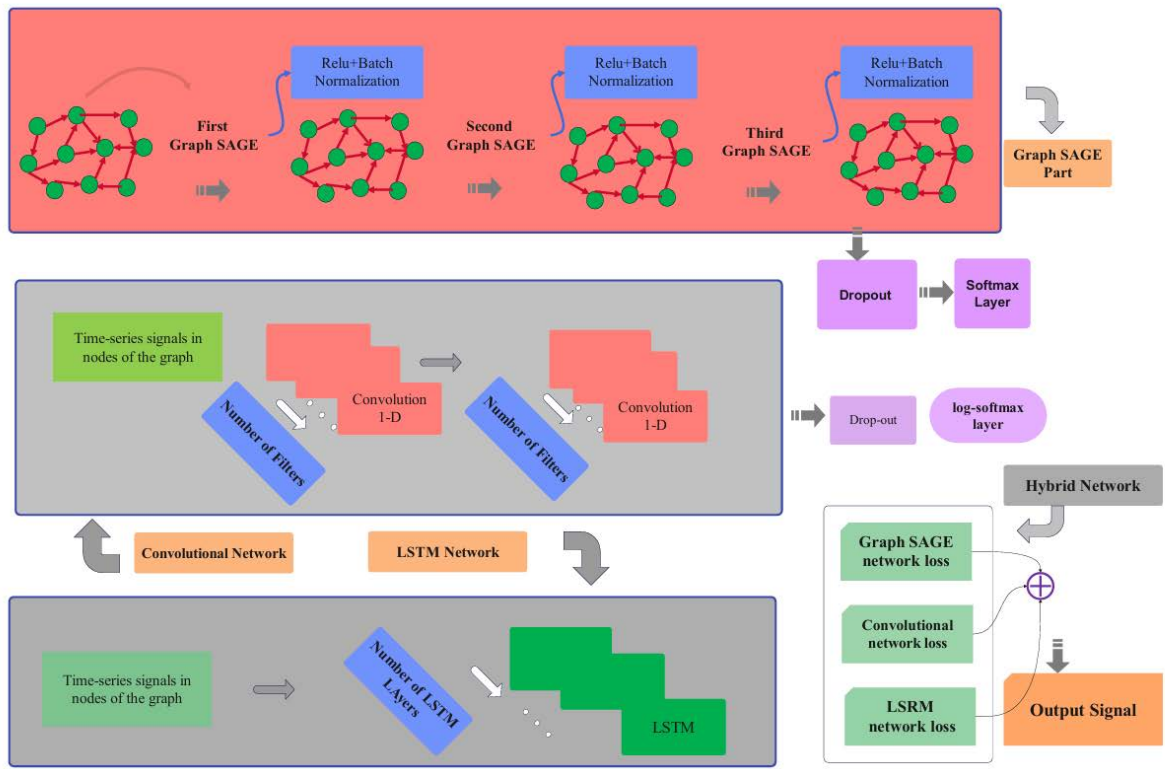

Figure 4. The detailed architecture of the proposed H-GSN.

The output of the pre-processing stage is imposed to the parallel convolutional part of the hybrid network. The loss function is the ensemble of two loss functions of the parallel parts of the H-GSN network. After log-softmax layers in parallel networks the obtained signal is classified according to the target vector. Batch normalization makes the network to be stable throughout training procedure and the convergence of the network would happen more quickly. The normalization is allocated to each graphsage layer. After four layers of graphsage and two parallel convolution and LSTM layers, the extracted feature array is acquired which is compatible with the size of target vector.

Table 7. Layers of the Graphsage of the suggested method.

| Layer | Layer Name | Activation Function | Shipping for Shipment Mode | | | Smart Logistics for Logistcs ID | | |
|---|---|---|---|---|---|---|---|---|
| | | | Dimension of Weight Array | Dimension of Bias | Number of parameters | Dimension of Weight Array | Dimension of Bias | Number of parameters |
| 1 | GraphSage | - | [1, 8, 8] | [8] | 72 | [1, 10,10] | [10] | 110 |
| 2 | Activation Layer | Relu | | | | | | |
| 3 | Batch normalization | - | [8] | [8] | 16 | [10] | [10] | 20 |
| 4 | GraphSage | - | [1,8,5] | [5] | 45 | [1, 10,10] | [10] | 110 |
| 5 | Activation Layer | Relu | | | | | | |
| 6 | Batch normalization | - | [5] | [5] | 10 | [10] | [10] | 20 |
| 7 | GraphSage | - | [1, 5, 3] | [3] | 18 | [1,10,10] | [20] | 110 |
| 8 | Activation Layer | Relu | | | | | | |
| 9 | Batch normalization | - | [3] | [3] | 6 | [10] | [10] | 20 |
| 10 | GraphSage | - | [1, 3, 3] | [3] | 12 | [1,10,10] | [10] | 110 |
| 11 | Activation Layer | Relu | | | | | | |
| 12 | Batch normalization | - | [3] | [3] | 6 | [10] | [10] | 10 |

The details and characteristics of the proposed architecture are explicated in Table 7 and 8. Table 7 is related with the details of graphsage part of the H-GSN. Tables 8 and 9 are the attributes of layers matching to the convolutional and LSTM parts of the network. Also, it shows the kernel size for different layers, the size of strides in layers, number of kernels used for each layer and the total number of weights to be trained during the training procedure.

The target vector for delivery status prediction in DataCo is a binary vector. The dimension of the target vector for the Shipping database is 5 for warehouses, 3 for shipping-mode classification and 2 for reaching-time classification. The dimension for the Smart Logistics classification is equal to 10 for logistics ID, 3 for traffic status and 3 for shipment status and a binary vector for logistics delay objective.

Table 8. Details of the convolutional part of the proposed method.

| Data | Layer | Layer Name | Activation Function | Output Dimension | Stride Shape | Size of Window | Number of Kernels | Number of Weights |
|---|---|---|---|---|---|---|---|---|
| Shipping (Logistic ID) | 1 | Convolution 1-D | LeakyReLU(alpha=0.1) | (10, 10, 5) | 1×1 | 1×5 | 10 | 510 |
| | 2 | Convolution 1-D | LeakyReLU(alpha=0.1) | (10, 10, 5) | 1×1 | 1×5 | 10 | 502 |
| Smart Logistics (Shipment Mode) | 3 | Convolution 1-D | LeakyReLU(alpha=0.1) | (8, 8, 5) | 1×1 | 1×5 | 8 | 328 |
| | 4 | Convolution 1-D | LeakyReLU(alpha=0.1) | (3, 8, 5) | 1×1 | 1×5 | 3 | 123 |

Table 9. Details of the LSTM part of the proposed method.

| Data | Layer | Layer Name | Number of Layers |
|---|---|---|---|
| Shipping (Logistic ID) | 1 | LSTM | 5 |
| | 2 | Linear | 1 |
| Smart Logistics (Shipment Mode) | 3 | LSTM | 5 |
| | 4 | Linear | 1 |

### 4.4. Training and evaluation of the proposed H-GSN:

In training procedure, the generated input and target samples are utilized to tune the parameters of the suggested H-GSN to the Dataco, Shipping and Smart Logistics datasets; we implement a 10-fold cross validation. After training and tuning the variables and parameters of the graphsage network and the parallel convolutional network, the testing phase is performed. The training of the proposed H-GSN is performed according to the parameters in Table 7-9. The optimal weights are attained and summarized in this table. The cross-validation is selected for the validation procedure.

**Algorithm 2: Pseudo-code for the proposed H-GSN.**

Proposed Hybrid GraphSage Network (H-GSN)

**Input:** 1) Data vectors $X$, 2) A threshold level , 3) Window size for adjacency matrix,

4) Number of layers for parallel parts of the hybrid network,

5) Labeled train and test samples $Xtrain$ and $Xtest$,

**Output:** Class Labels for $X_{test}$

Initialization of the parameters regarding the model.

Training corresponding to the 10-fold cross-validation:

1: Determine the correlation co-efficient of the of $X$ in $Xtrain$.

2: Calculate the adjacency matrix $W$ via using sigmoid function for the result of step 1.

3: Extract the output of the graphSAGE layers.

6: Calculate the output of the dropout layer.

7: Calculate the output of the parallel convolutional and LSTM layers.

8: Multi-Objective optimization of the weights of the hybrid layers using optimal loss function.

9: Update the weights of the layers regarding the totals hybrid cost function:

$$\begin{cases} Loss_{Cross-Entropy}(t\arg et, output_{1GraphSage}) = -\frac{1}{n}\sum_{i=1}^{n}(t\arg et_i.\log output_{1i} + (output_{1i} - t\arg et_i).\log(t\arg et_i - real_i)) \\ Loss_{Cross-Entropy}(t\arg et, output_{2Convolution}) = -\frac{1}{n}\sum_{i=1}^{n}(t\arg et_i.\log output_{2i} + (output_{2i} - t\arg et_i).\log(t\arg et_i - real_i)) \\ Loss_{Cross-Entropy}(t\arg et, output_{3LSTM}) = -\frac{1}{n}\sum_{i=1}^{n}(t\arg et_i.\log output_{3i} + (output_{3i} - t\arg et_i).\log(t\arg et_i - real_i)) \end{cases}$$

$$Loss_{Total} = Loss_{Cross-Entropy}(t\arg et, output_1) + Loss_{Cross-Entropy}(t\arg et, output_2) + Loss_{Cross-Entropy}(t\arg et, output_3)$$

10: Attain the predictions for the graph illustrations in accordance with $Xtest$ using the trained H-GSN.

Stop specifications: A maximum number of trials or acceptable accuracy.

A 10-fold cross-validation is fulfilled using the training samples. The test stage can predict the multi-objective targets for 3 databases based on the calculated weights of the training stage. The pseudo-code in algorithm 1 explains the details of the proposed H-GSN. Table 6 shows the considered training search area and the optimal parameters for each scope.

Table 10. Details of training variables.

| Parameters | Search Scope | Optimal Value |
|---|---|---|
| Optimizer of GraphSage | Adam, SGD | Adam |
| Cost function of GraphSage | MSE, Cross-Entropy | Cross-Entropy |
| Number of Sage layers | 2, 3, 4 | 4 |
| Learning rate of GraphSage | 0.1, 0.01, 0.001 | 0.001 |
| Window size | 10, 20, 30 | 20 |
| Optimizer of convolution and LSTM | Adam, SGD | Adam |
| Learning rate of convolution and LSTM | 0.01, 0.001, 0.0001, 0.00001 | 0.0001 |
| Number of convolution layers | 2, 3, 4 | 2 |

## 5. Results and Discussion

In this section, the obtained results of analysis of the proposed H-GSN are presented. Our configuration is executed on a laptop with a 16 GB RAM, 2.8 GHz Core i7 CPU and a GeForce GTX 1050 GPU. The implementation of the proposed network is performed using the Google Colaboratory Pro platform.

Figure 5 shows the performance of proposed H-GSN and H-GatN for DataCo based on the accuracy corresponding to prediction of 4 different shipment modes. The graph attention is used in the H-GatN instead of the graphsage layers. Corresponding to this figure, Adam optimizer with optimal learning rate of 0.0001 and optimum weight decay of $4 \times (10^{-4})$ have been used taking into consideration the cross-entropy for the first segment of the network and the total loss corresponding to the pseudocode for the ensemble segment of the proposed network. This figure illustrates the accuracy plots for H-GSN, H-GatN, GSN and GatN. Regarding same number of iterations corresponding to four various methods, the proposed hybrid graphsage demonstrates better performance. The graphsage and graph attentional methods have a weak performance in comparison to the hybrid ones. Four layers of graph attention networks have been considered for H-GatN and GatN. As it can be seen, we consider more than 400 number of iterations for all methods utilizing a 10-fold cross-validation.

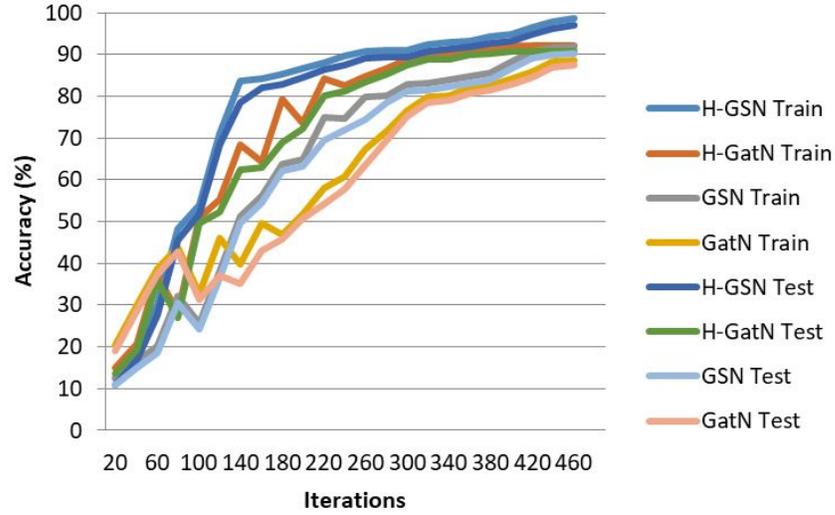

Figure 5. Accuracy plots for training the DataCo corresponding to Shipping Mode classification scenario.

Table 11 reports the performance metrics considering the DataCo for prediction of the delivery status and shipment type for different methods. This table shows the on-time delivery and late delivery status prediction accuracy. Besides, it demonstrates the precision, F1-score and recall considering various orders for hybrid graphsage network, hybrid graph attention, non-hybrid graphsage and non-hybrid graph attention methods.

Table 11. Performance Metrics of the proposed method(Accuracy, Precision, Recall, F1-score) regarding DataCo.

| DataCo Categories (Logistic Delay) | H-GSN | GSN | H-GatN | GatN | DataCo Categories (Shipping Mode) | H-GSN | GSN | H-GatN | GatN |
|---|---|---|---|---|---|---|---|---|---|
| **Overall accuracy** | 99.9 | 92.56 | 94.98 | 90.45 | **Overall accuracy** | 98.7 | 91.82 | 92.15 | 88.65 |
| **Precision** | 99.9 | 92.5 | 94.9 | 90.74 | **Precision** | 98.7 | 91.8 | 91.1 | 86.6 |
| **F1-score** | 98.9 | 92.5 | 94.9 | 90.42 | **F1-score** | 98.7 | 91.8 | 92.1 | 86.6 |
| **Recall** | 99.9 | 92.5 | 94.9 | 90.08 | **Recall** | 98.7 | 91.8 | 91.1 | 86.6 |

Figures 6 and 7 illustrate the two dimensional T-SNE plots for different layers of the proposed H-GSN in order to demonstrate the procedure of the classification and a tangible view of stages of classification considering the proposed H-GSN with DataCo dataset.

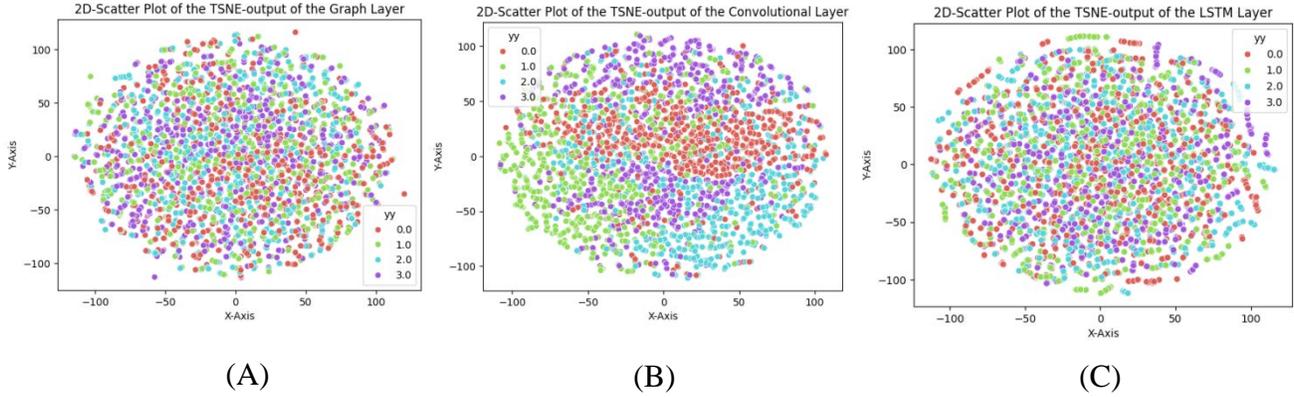

(A)  (B)  (C)

Figure 6. Two-dimensional TSNE plots for DataCo. (A) Graph layer, (B) Convolution layer (C) LSTM layer.

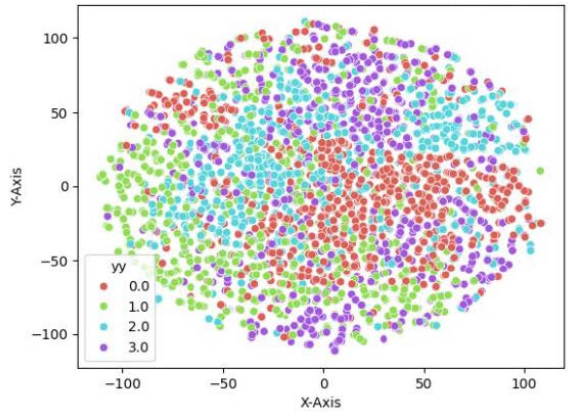

Figure 7. Two-dimensional TSNE plots for DataCo of the hybrid output.

Table 12 is the report for the classification objectives regarding the Shipping database. The classification accuracies for different types of warehouses 'A', 'B', 'C', 'D' and 'F' can be seen in this table. Moreover, the results of the proposed H-GSN together with H-GatN, GSN and GatN are available in this table for categorization purposes of shipment modes of 'flight', 'ship', and 'road'. This table presents the binary classification results regarding reaching time in logistics for supply chain. As it can be seen in Table 12, the proposed H-GSN surpasses the other methods.

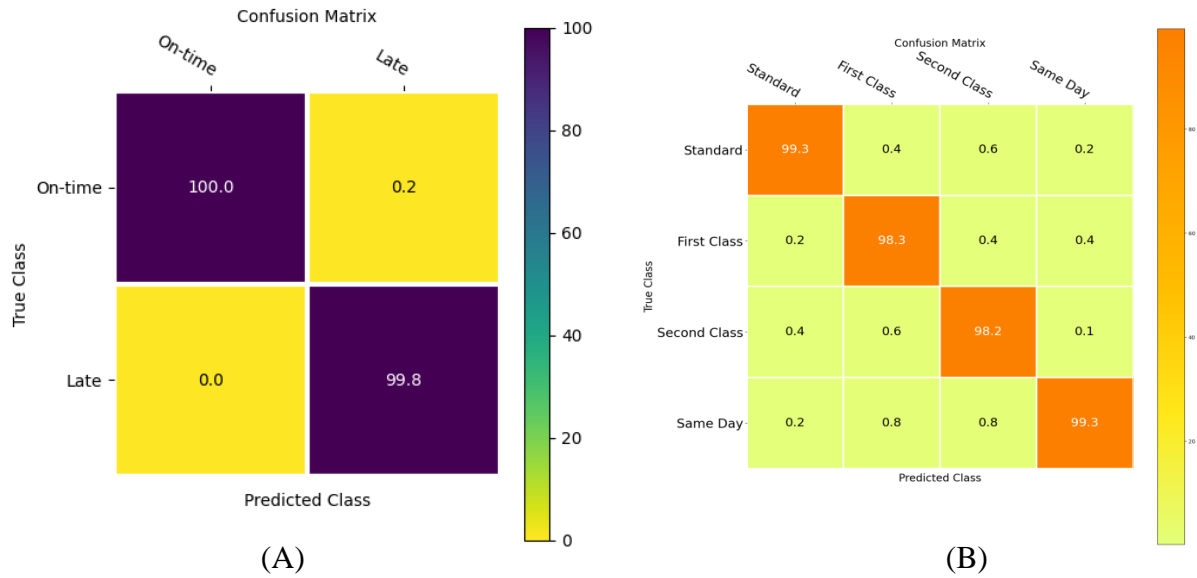

Figure 8. Confusion matrix for DataCo regarding (A) Delivery Risk (B) Shipping Mode.

Table 12. Performance Metrics of the proposed method(Accuracy, Precision, Recall, F1-score) in Shipping database.

| Shipping Database (Reached Time) | H-GSN | H-GatN | Shipping Database (Mode of Shipment) | H-GSN | H-GatN | Shipping Database (Warehouses) | H-GSN | H-GatN |
|---|---|---|---|---|---|---|---|---|
| **Overall accuracy** | 99.4 | 96.54 | **Overall accuracy** | 90.33 | 85.46 | **Overall accuracy** | 100 | 97.28 |
| **Precision** | 99.4 | 96.18 | **Precision** | 90.3 | 85.4 | **Precision** | 100 | 97.2 |
| **F1-score** | 99.4 | 96.03 | **F1-score** | 90.3 | 85.4 | **F1-score** | 100 | 97.2 |
| **Recall** | 99.4 | 96.19 | **Recall** | 90.3 | 85.4 | **Recall** | 100 | 97.2 |

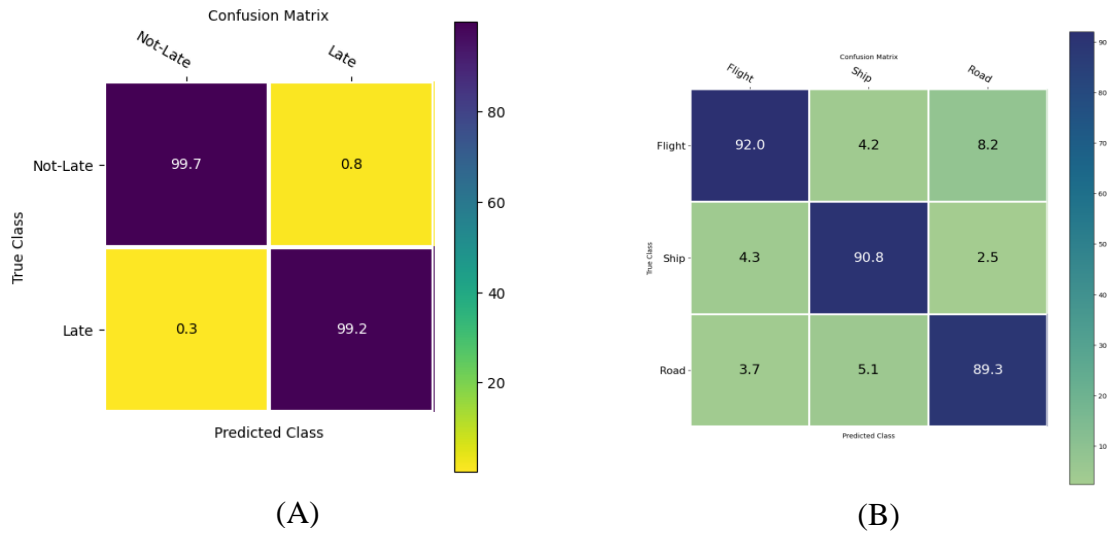

(A)                  (B)

Figure 9. The confusion matrix for Shipping Database regarding (A) Reached Time (B) Mode of Shipment.

The confusion matrix is valuable way of confirming the efficiency of the proposed method. Figure 8 delineates the confusion matrix for DataCo dataset regarding delivery status and shipment mode prediction of 'standard', 'first class', 'second class' and 'same day'. Figure 9 shows the performance of the proposed H-GSN considering the Shipping dataset. Figure 10 is the confusion matrix for the classification of the warehouse types considering our proposed H-GSN.

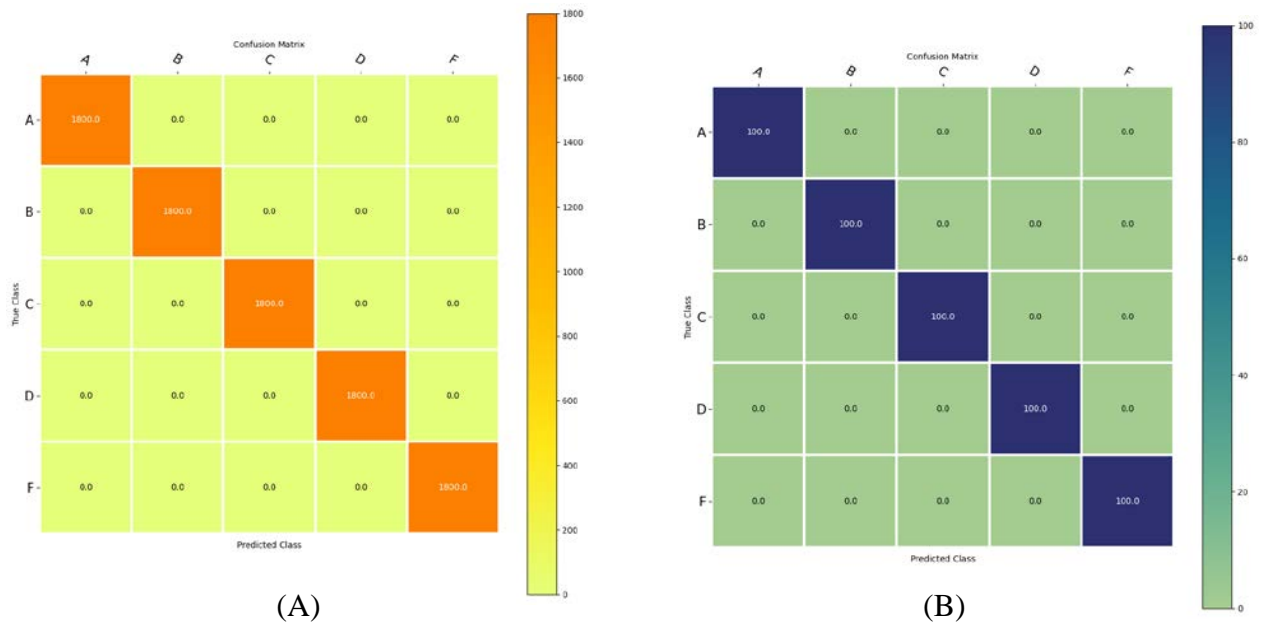

(A)                  (B)

Figure 10. Product Classification for Shipping Database regarding Warehouses (A) in Number (B) in Percent.

Table 13 shows the performance metrics of the proposed method in comparison to the other methods for Smart Logistics database. As it can be seen, our proposed geometric hybrid graphsage network outperforms the other hybrid graph attention method.

Table 13. Accuracy for multi-task classification of the Smart Logistics database.

| Smart Logistics (Logistics ID) | H-GSN | H-GatN | Smart Logistics (Shipment Status) | H-GSN | H-GatN | Smart Logistics (Logistics Delay) | H-GSN | H-GatN |
|---|---|---|---|---|---|---|---|---|
| **Overall accuracy** | 97.8 | 88.46 | **Overall accuracy** | 100 | 90.4 | **Overall accuracy** | 96.35 | 80.2 |
| **Precision** | 97.8 | 87.18 | **Precision** | 100 | 89.3 | **Precision** | 96.38 | 80.2 |
| **F1-score** | 97.8 | 87.03 | **F1-score** | 100 | 89.2 | **F1-score** | 96.32 | 80.2 |
| **Recall** | 97.8 | 87.19 | **Recall** | 100 | 89.26 | **Recall** | 96.35 | 80.2 |

Figure 11 is the confusion matrixes for classification of the Smart Logistics considering the proposed H-GSN. Figure 11(A) considers logistics delay according to the Smart Logistics database. Figure 11(B) considers the shipment status as the target task for this database. All of them confirm the efficiency of the proposed method.

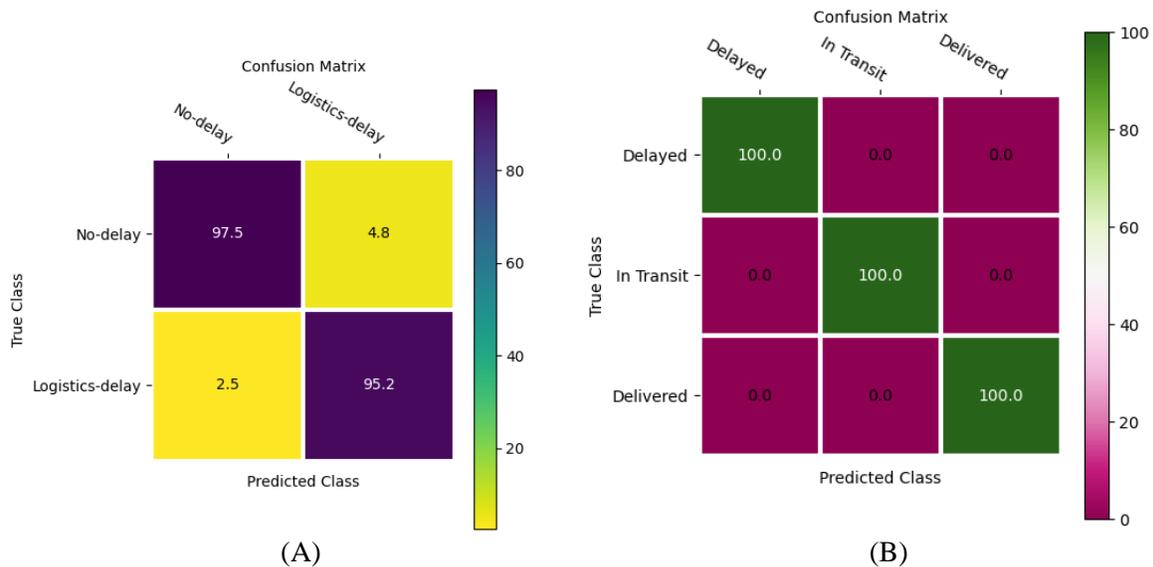

(A)  (B)

Figure11. The confusion matrix for Smart Logistics Database regarding different scenarios: (A)Logistics Delay, (B)Shipment Status.

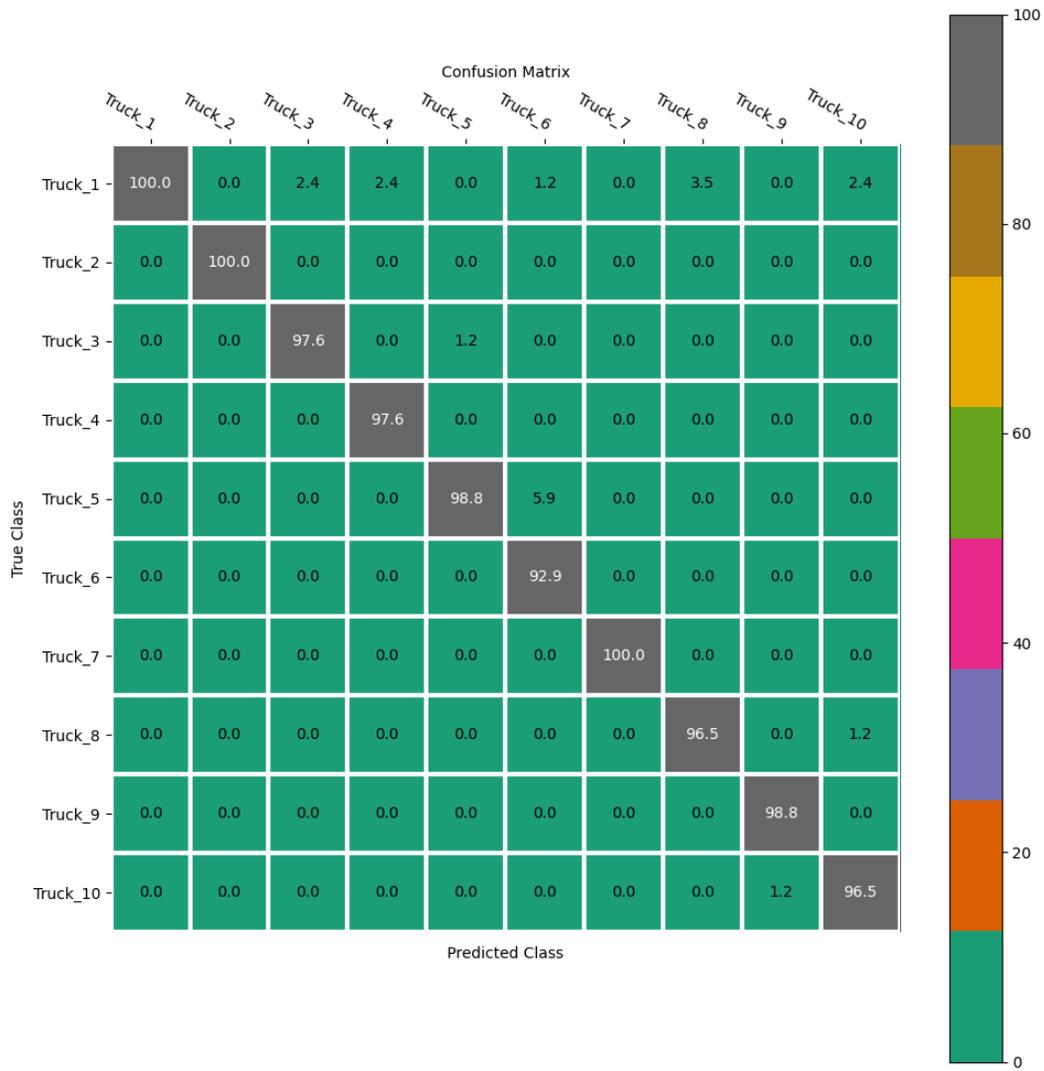

Figure 12. The confusion matrix for Smart Logistics Database regarding Logistics ID (Truck ID).

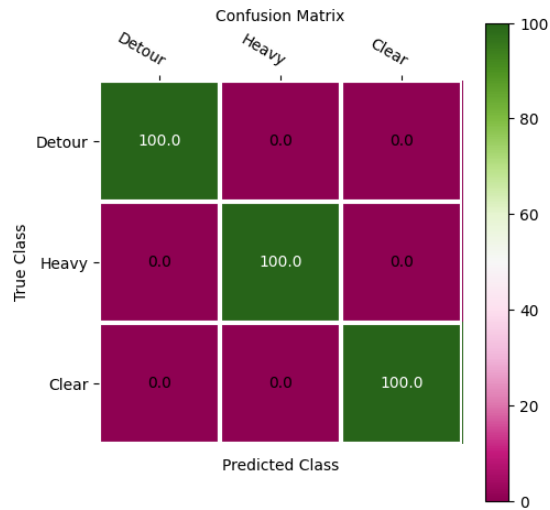

Figure 13. The confusion matrix for Smart Logistic Database regarding Traffic Status.

Figures 12 and 13 showcase the confusion matrix for logistics ID and traffic status classification of the Smart Logistics dataset, respectively. Table 14 shows the performance metrics of the proposed method in comparison to the other novel and traditional methods. As it can be seen, our proposed geometric hybrid network outperforms the other conventional methods.

Table 14. Comparison with the conventional methods.

| Method | Logistic ID Smart Logistics Database | Shipment Status Smart Logistics Database | Logistic Delay Smart Logistics Database | Traffic Status Smart Logistics Database |
|---|---|---|---|---|
| H-GSN | 97.8 | 100 | 92.37 | 100 |
| GNN-based [35] | 81.23 | 92.36 | 93.46 | 91.82 |
| KNN [36] | 63.44 | 78.23 | 79.65 | 76.43 |
| XGB [37] | 62.42 | 74.06 | 76.04 | 73.15 |
| Lgistic regression | 66.67 | 68.32 | 69.54 | 68.21 |

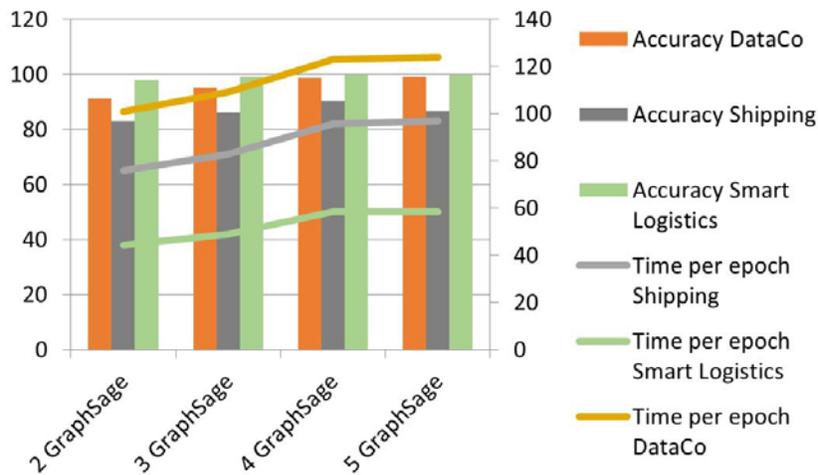

Figure 14. Accuracy and time of training per epoch with different number of graphSage layers for Shipment Mode scenario.

To investigate the effect of different parameters on the optimality of the performance, we execute an extended experiment. In order to evaluate the effect of alternating the number of sequential graphsage layers, a series of training procedures are done for different numbers of sequential graphsage layers. Figure 14 showcases the results of tuning for 2, 3, 4 and 5 sequential graphsage layers. Setting the sequential layers more than four in this case study does not improve the performance, it affect the computational complexity. This figure showcases the incremental direction of the training time per iteration epoch of the proposed H-GSN.

Table 14 provides performance metrics for the proposed H-GSN in comparison to other conventional methods considering different classification scenarios regarding the Smart Logistics database.

## 6. Conclusion

In this paper, a novel deep hybrid geometric architecture is proposed to solve the problem of logistics, logistics risk management in a supply chain. In addition, it is a deep model proposed to strengthening the sustainability of a supply chain. The proposed model architecture is used for testing the sustainability of the DataCo, Shipping and Smart Logistics databases and it is used for logistics management regarding the mentioned three databases.

The main challenges in this paper are utilizing the graph theory along with deep network architectures considering the connectivity between nodes to extract the hidden states of supply chain principle vectors. The proposed hybrid geometric deep network is a novel approach for multi-objective classification. It is a multi-task network which facilitates the logistics management in a supply chain along with strengthening the sustainability of a supply chain. The efficiency of the proposed method is delineated via the exploratory outcomes on three supply chain logistics datasets available on Kaggle website. Comparing the results with other novel state-of-the-art methods confirms that the H-GSN receives higher classification accuracy for node and edge classification with fewer numbers of iterations.